\documentclass[11pt]{article}
\usepackage{coling2020}
\usepackage{times}
\usepackage{url}
\usepackage{latexsym}
\usepackage{hyperref}
\usepackage{graphicx}
\usepackage{multirow}
\usepackage{booktabs}

\colingfinalcopy 

\title{A Multi-Modal Method for Satire Detection using Textual and Visual Cues}

\author{Lily Li$^1$, Or Levi$^2$, Pedram Hosseini$^3$, David A. Broniatowski$^3$ \\
  $^1$Jericho Senior High School, New York, USA \\
  $^2$AdVerifai, Amsterdam, Netherlands \\
  $^3$The George Washington University, Washington D.C., USA \\
  {\tt lily.li@jerichoapps.org, or@adverifai.com} \\{\tt \{phosseini,broniatowski\}@gwu.edu}
  }

\date{}

\begin{document}
\maketitle

\begin{abstract}
Satire is a form of humorous critique, but it is sometimes misinterpreted by readers as legitimate news, which can lead to harmful consequences. We observe that the images used in satirical news articles often contain absurd or ridiculous content and that image manipulation is used to create fictional scenarios. While previous work have studied text-based methods, in this work we propose a multi-modal approach based on state-of-the-art visiolinguistic model ViLBERT. To this end, we create a new dataset consisting of images and headlines of regular and satirical news for the task of satire detection. We fine-tune ViLBERT on the dataset and train a convolutional neural network that uses an image forensics technique. Evaluation on the dataset shows that our proposed multi-modal approach outperforms image-only, text-only, and simple fusion baselines.

\end{abstract}

\section{Introduction}

Satire is a literary device that writers employ to mock or ridicule a person, group, or ideology by passing judgment on them for a cultural transgression or poor social behavior. Satirical news utilizes humor and irony by placing the target of the criticism into a ridiculous, fictional situation that the reader must suspend their disbelief and go along with \cite{maslo2019parsing}. However, despite what absurd content satirical news may contain, it is often mistaken by readers as real, legitimate news, which may then lead to the unintentional spread of misinformation. In a recent survey conducted by The Conversation\footnote{https://theconversation.com/too-many-people-think-satirical-news-is-real-121666}, up to 28\% of Republican respondents and 14\% of Democratic respondents reported that they believed stories fabricated by the Babylon Bee, a satirical news website, to be “definitely true”. In these instances, the consequences of satire are indistinguishable from those of fake news.

\begin{figure*}[h!]
    \centering
    \includegraphics[scale=0.21]{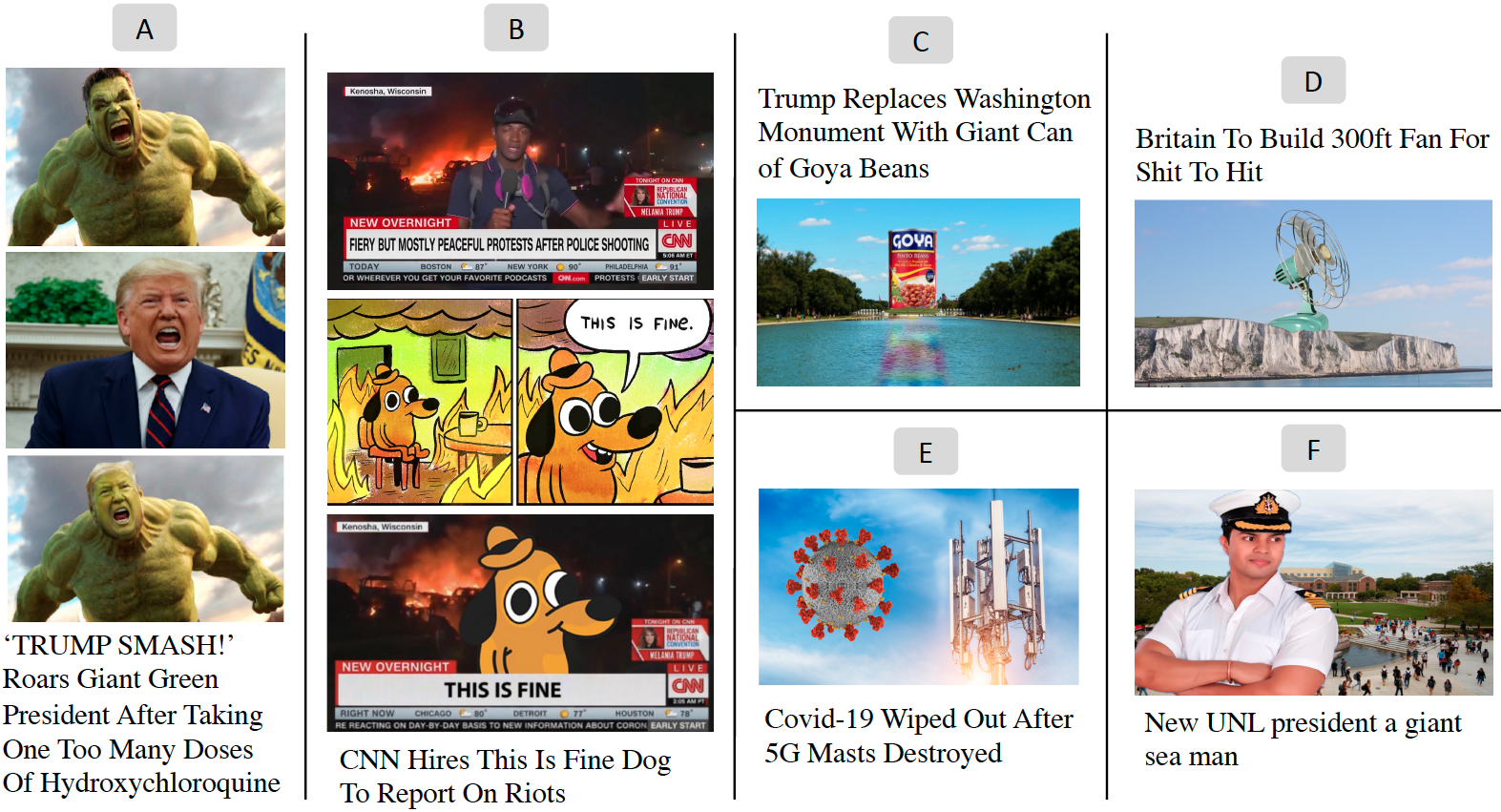}
    \caption{Examples of satirical news images created by altering existing images.}
    \label{fig:my_label}
\end{figure*}

To reduce the spread of misinformation, social media platforms have partnered with third-party fact-checkers to flag false news articles and tag articles from known satirical websites as satire for users \cite{facebook,google}. However, due to the high cost and relative inefficiency of employing experts to manually annotate articles, many researchers have tackled the challenge of automated satire detection. Existing models for satirical news detection have yet to explore the visual domain of satire, even though image thumbnails of news articles may convey information that reveals or disproves the satirical nature of the articles. In the field of cognitive-linguistics, \newcite{maslo2019parsing} observed the use of altered images showing imaginary scenarios on the satirical news show The Daily Show. This phenomenon also extends to satirical news articles, as seen in Figure 1. For example, Figure 1(a) depicts the Marvel Cinematic Universe character Hulk from the film Avengers: Infinity War and the United States President Donald Trump spliced together. Alone, each of the two images is serious and not satirical, but, since they come from drastically different contexts, combining the two images creates a clearly ridiculous thumbnail that complements the headline of the article.

In our work, we propose a multi-modal method for detecting satirical news articles. We hypothesize that 1) the content of news thumbnail images when combined with text, and 2) detecting the presence of manipulated or added characters and objects, can aid in the identification of satirical articles.

\section{Related Work}
Previous work proposed methods for satirical news detection using textual content \cite{levi2019}. Some works utilize classical machine learning algorithms such as SVM with handcrafted features from factual and satirical news headlines and body text, including bag-of-words, n-grams, and lexical features \cite{burfoot-baldwin-2009-automatic,rubin-etal-2016-fake}. More recent works use deep learning to extract learned features for satire detection. \newcite{yang-etal-2017-satirical} proposed a hierarchical model with attention mechanism and handcrafted linguistic features to understand satire at a paragraph and article-level.

While previous work utilize visiolinguistic data for similar tasks, there is no related work that employs multi-modal data to classify articles into satirical and factual news. \newcite{nakamura2019rfakeddit} created a dataset containing images and text for fake news detection in posts on the social media website Reddit. While they include a category for satire/parody in their 6-way dataset, since they use only content that has been submitted by Reddit users, it is not representative of mainstream news media. Multi-modal approaches have also been tried in sarcasm detection; \newcite{castro-2019} compiled a dataset of scenes from popular TV shows and \newcite{cai-etal-2019-multi} used tweets comprising of text and images from Twitter.


\section{Methods}

\subsection{Data}
We create a new multi-modal dataset of satirical and regular news articles. 
The satirical news is collected from four websites that explicitly declare themselves to be satire, and
the regular news is collected from six mainstream news websites\footnote{The regular news websites we use are listed by Media Bias/Fact Check \url{https://mediabiasfactcheck.com/}, a volunteer-run and nonpartisan organization dedicated to fact-checking and determining the bias of news publications}. 
Specifically, the satirical news websites we collect articles from are The Babylon Bee, Clickhole, Waterford Whisper News, and The DailyER. The regular news websites are Reuters, The Hill, Politico, New York Post, Huffington Post, and Vice News. We collect the headlines and the thumbnail images of the latest 1000 articles for each of the publications. The dataset contains a total of 4000 satirical and 6000 regular news articles.

\subsection{Proposed Models}
\textbf{Multi-Modal Learning.} We use Vision \& Language BERT (ViLBERT), a multi-modal model proposed by \newcite{lu2019vilbert} that processes images and text in two separate streams. Each stream consists of transformer blocks based on BERT \cite{devlin2018bert} and co-attentive layers that facilitate interaction between the visual and textual modalities. In each co-attentive transformer layer, multi-head attention is computed the same as a standard transformer block except the visual modality attends to the textual modality and vice-versa. To learn representations for vision-and-language tasks, ViLBERT is pre-trained using the masked multi-model modeling and multi-modal alignment prediction tasks on the Conceptual Captions dataset \cite{sharma-etal-2018-conceptual}. We choose to use ViLBERT because of its high performance on a variety of visiolinguistic tasks, including Visual Question Answering, Image Retrieval, and Visual Commonsense Reasoning. We fine-tune ViLBERT on the satire detection dataset by passing the element-wise product of the final image and text representations into a learned classification layer.

\textbf{Image Forgery Detection.} Since satirical news images are often forged from two or more images (known as image splicing), we implement an additional model that uses error level analysis (ELA). ELA is an image forensics technique that takes advantage of lossy JPEG compression for image tampering detection \cite{krawetz2007picture}. In ELA, each JPEG image is resaved at a known compression rate, and the absolute pixel-by-pixel differences between the original and the resaved images are compared. 
ELA can be used to identify image manipulations where a lower quality image was spliced into a higher quality image or vice-versa. To detect image forgeries as an indicator of satirical news, we preprocess the images using ELA with a compression rate of 90\% and use them as input into a CNN.

For the CNN, we use two convolutional layers with 32 kernels and a filter width of 5, each followed by a max-pooling layer. The output features from the CNN are fed into a MLP with a hidden size of 256 and a classification layer. We pretrain the model on the CASIA 2.0 image tampering detection dataset \cite{dong2013casia} before fine-tuning on the images of the satire detection dataset.

\textbf{Implemention.} We divide the data into training and test sets with a ratio of 80\%:20\%. We train all our models with a batch size of 32 and Adam optimizer. We use the MMF \cite{singh2020mmf} implementation of ViLBERT and fine-tune it for 12 epochs with a learning rate of 5e-6. We extract Mask RCNN \cite{He_2017} features from the images in the dataset as visual input. The ViLBERT model has 6 transformer blocks in the visual stream and 12 transformer blocks in the textual stream. Our ELA+CNN model is trained with a learning rate of 1e-5 for 7 epochs.\footnote{Scripts for our experiments are available at: \url{https://github.com/lilyli2004/satire}}

\subsection{Baselines}
To create fair baselines for our fine-tuned ViLBERT model, we train multi-modal models that use simple fusion. In the model denoted as Concatenation, ResNet-101 \cite{he2016resnet} and BERT features are concatenated and a MLP is trained on top. In the model denoted as Average fusion, the output of ResNet-101 and BERT are averaged. We choose these two models as our baselines to evaluate the effects of ViLBERT's early fusion of visual and textual representations and multi-modal pre-training on Conceptual Captions \cite{sharma-etal-2018-conceptual}. We also fine-tune uni-modal ResNet-101 and BERT\textsubscript{BASE} models to compare the performance of the multi-modal models to.

\begin{table*}[th]
    \small
    \def\arraystretch{1.3}
    \setlength{\belowcaptionskip}{-15pt}
    \centering
    \begin{tabular}{ccccc}
    \bottomrule
    Type & Model & Accuracy & F1 score & AUC-ROC \\ \bottomrule
     & All regular news & 60.00 & — & 50.00 \\ \hline
    \multirow{4}{*}{Baselines} & ResNet101 & 73.54 & 65.26 & 80.28 \\
     & BERT\textsubscript{BASE} & 91.33 & 88.64 & 96.77 \\
     & Simple fusion (average) & 92.53 & 90.44 & 96.74 \\
     & Simple fusion (concatenation) & 92.74 & 90.70 & 97.31 \\ \hline
    \multirow{2}{*}{Proposed Models} & ELA+CNN & 44.20 & 51.86 & 44.61 \\
     & ViLBERT & \textbf{93.80} & \textbf{92.16} & \textbf{98.03} \\ \bottomrule
    \end{tabular}
    \caption{Model performance on satire detection dataset.}
    \label{tab:my-table}
\end{table*}

\section{Results and Discussion}

\subsection{Experimental Results}
We measure the performance of the proposed and baseline models using Accuracy, F1 score, and AUC-ROC metrics. The results are shown in Table 1. The models using only the visual modality (ResNet-101 and CNN+ELA) do not perform as well as the model that uses only the text modality (BERT\textsubscript{BASE}). The simple fusion models (Average fusion, Concatenation) perform marginally better than BERT\textsubscript{BASE}. 

Surprisingly, the performance of the ELA+CNN model was very poor, achieving an accuracy worse than random chance. While this is not in line with our initial hypothesis, there might be several reasons for these results: Firstly, ELA is not able to detect image manipulations if the images have been resaved multiple times since after they have been compressed at a high rate there is little visible change in error levels \cite{krawetz2007picture}. This makes it especially difficult to identify manipulation in images taken from the Internet, as they have usually undergone multiple resaves and are not camera originals. Additionally, although ELA can be used as a method to detect and localize the region of an image that has been potentially altered, it does not allow for the identification of what kind of image manipulation technique was used. This is important because even reputable news publications, such as Reuters and The Associated Press use Photoshop and other software to perform minor adjustments to photos, for example, to alter the coloring or lighting, or to blur the background \cite{reutersphotoshop,associatedpress}. 

ViLBERT outperforms the simple fusion multi-modal models because it uses early, deep fusion and has undergone multi-modal pre-training rather than only separate uni-modal visual and text pre-training. ViLBERT also performs almost 3.5 F1 points above the uni-modal BERT\textsubscript{BASE} model.  

\subsection{Model Misclassification Study}

After classification, we randomly select 20\% of the test set samples misclassified by ViLBERT and observed them for patterns across multiple samples. Figure 2 shows examples of misclassified samples. We observed three main reasons that may have been the cause of the incorrectly classified articles: 1) The model misinterpreted the headline (Figure 2(a)), 2) the model lacks knowledge of current events (Figure 2(b)), and 3) the article covered a bizarre but true story (Figure 2(c)).

Figure 2(a) shows an article from Politico that has been classified as satire. The image does not portray anything strange or out of the ordinary. However, the headline uses the word ``bursts", which the model might be incorrectly interpreting in the literal sense even though it is being used metaphorically. If ``bursts" was intended to be literal, it would drastically change the meaning of the text, which may be why the model failed to classify the article as factual.
Figure 2(b) shows a satirical article from Babylon Bee that has been misclassified as factual. Its image has also not been heavily altered or faked; in fact, it is the same image that was used as the original thumbnail of the Joe Rogan podcast episode that is the subject of the article. However, the model fails to recognize the ridiculousness of the text, since it does not have the political knowledge to spot the contrast between the ``alt-right" and the American politician Bernie Sanders.
In Figure 2(c), an article is from the factual publication The New York Post is misclassified as satirical. Although both the headline and the image seem very ridiculous, the story and the image were, in fact, not fabricated. Thus, identifying text/images as absurd might not always aid in satire detection, since ViLBERT fails in classifying this article as factual because it is unable to tell that the image has not been forged.

\begin{figure*}[h!]
    \setlength\abovecaptionskip{0\baselineskip}
    \setlength{\belowcaptionskip}{-12pt}
    \centering
    \includegraphics[scale=0.55]{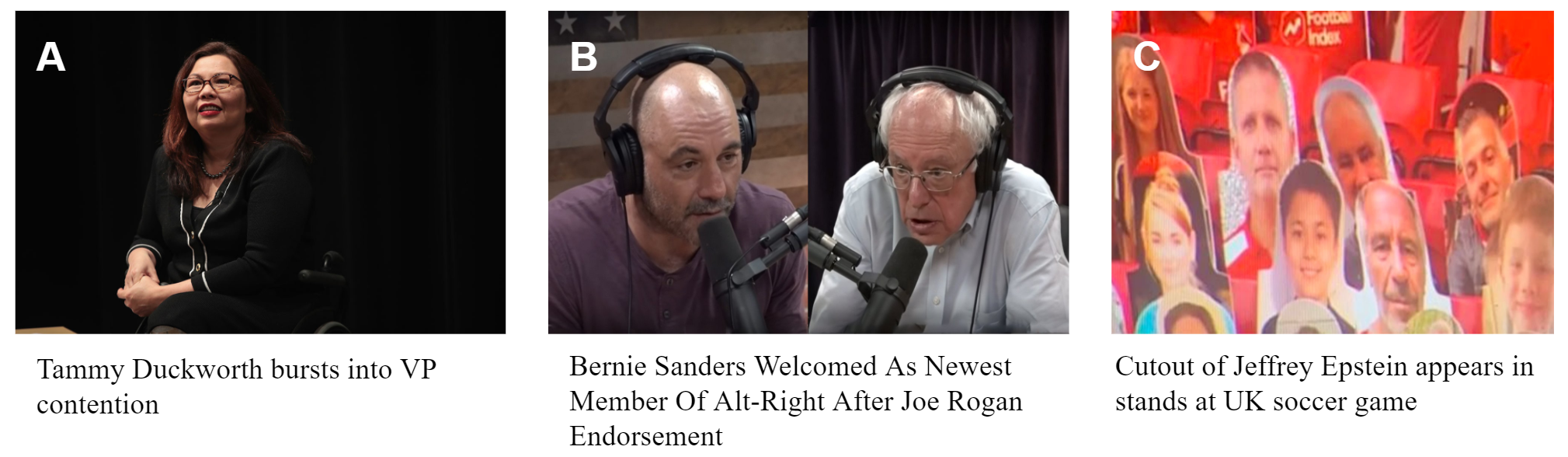}
    \caption{Examples of articles misclassified by ViLBERT}
    \label{fig:my_label}
\end{figure*}

\section{Conclusion and Future Investigations}
In this paper we create a multi-modal satire detection dataset and propose two models for the task based on the characteristics of satirical images and their relationships with the headlines. While our model based on image tampering detection performed significantly worse than the baselines, empirical evaluation showed the efficacy of our proposed multi-modal approach compared to simple fusion and uni-modal models. In future work on satire detection, we will incorporate image forensics methods to identify image splicing in satirical images, as well as knowledge about politics and other current issues.

%
%
\blfootnote{
    %
    %
    %
    %
    %
    
    \hspace{-0.65cm}  
    This work is licensed under a Creative Commons 
    Attribution 4.0 International License.
    License details:
    \url{http://creativecommons.org/licenses/by/4.0/}.
}

\bibliographystyle{coling}
\bibliography{coling2020}

\end{document}